\documentclass[
]{ceurart}

\usepackage{listings}
\usepackage{todonotes}
\usepackage{enumitem}

\newcommand{\lkgb}{LLM-KG-Bench\xspace}
\newcommand{\LKGBf}{LLM-KG-Bench framework\xspace} 
\newcommand{\sparqlselquery}{\emph{SPARQL SELECT} query\xspace}

\begin{document}

\begin{picture}(0,0)
  \put( \dimexpr(\textwidth+7mm), -0.5\textheight){%
      \rotatebox[origin=c]{90}{%
        \LARGE peer reviewed publication at \href{https://2025-eu.semantics.cc/page/posters}{SEMANTiCS 2025 Poster Track}
  }}
\end{picture}

\copyrightyear{2025}
\copyrightclause{Copyright for this paper by its authors.
  Use permitted under Creative Commons License Attribution 4.0
  International (CC BY 4.0).}

\conference{Posters \& Demos Track of SEMANTiCS’25:
International Conference on Semantic Systems, September 3–5, 2025, Vienna, Austria}
\title{Characterizing Knowledge Graph Tasks in LLM Benchmarks Using Cognitive Complexity Frameworks}

\author[1]{Sara Todorovikj}[%
orcid=0000-0002-2418-1358,
email=sara.todorovikj@informatik.tu-chemnitz.de
]
\cormark[1]
\address[1]{Chemnitz University of Technology, Germany}

\author[1,2]{Lars-Peter Meyer}[%
orcid=0000-0001-5260-5181
]
\address[2]{InfAI, Leipzig, Germany}

\author[1,2]{Michael Martin}[%
orcid=0000-0003-0762-8688
]

\cortext[1]{Corresponding author.}

\begin{abstract}
Large Language Models (LLMs) are increasingly used for tasks involving Knowledge Graphs (KGs), whose evaluation typically focuses on accuracy and output correctness. We propose a complementary task characterization approach using three complexity frameworks from cognitive psychology. Applying this to the \LKGBf, we highlight value distributions, identify underrepresented demands and motivate richer interpretation and diversity for benchmark evaluation tasks.
\end{abstract}

\begin{keywords}
   Task characterization \sep Benchmark evaluation \sep LLM \sep Knowledge Graph \sep RDF \sep SPARQL 
\end{keywords}

\maketitle

\section{Introduction}
Large Language Models (LLMs) are increasingly applied to structured knowledge tasks involving query generation, data interpretation and interaction with Knowledge Graphs (KGs) \cite{Pan2024UnifyingLargeLanguage,Meyer2023LLMassistedKnowledge}. As a result, the LLM-KG-Bench has been introduced \cite{Meyer2023DevelopingScalableBenchmark,Meyer2025LLMKGBench3} that aim to assess model performance in KG-related context in an automated way. Evaluation for such benchmark tasks typically focuses on correctness and surface-level output features, but provide limited insight into the deeper nature of the tasks themselves, specifically, what kinds of knowledge and operations they demand. In this paper, we propose a task characterization framework for evaluation benchmarks using cognitive complexity frameworks. These allow us to describe each task in terms of the minimal operational and structural requirements expected for successful task completion.
Our aim is to support deeper understanding of task diversity and complexity and complement performance-oriented evaluation with structure insight, extending our previous work \cite{Meyer2025EvaluatingLLMsForRdf}.

\section{Background and Related Work}
Understanding the difficulty and structure of tasks often requires going beyond surface-level features. In cognitive science and educational research, several frameworks have been developed to describe the complexity of tasks based on the type of knowledge involved and the mental operations required. One of the most well-known is \emph{Bloom's Taxonomy} \cite{bloom1956}, which was originally developed for classification of educational goals based on the required cognitive complexity level. The taxonomy is grounded on behavioral observations of learning processes and classifies cognitive processes from simple recall to higher-level reasoning and creative generation. A revision was made in order to better fit modern views of cognitive psychology \cite{anderson2001}, which also introduced a complementary dimension. The new \emph{Knowledge Dimension} distinguishes between types of knowledge required for completing different tasks. In parallel, \emph{Relational Complexity Theory} \cite{halford1998} originates from developmental and comparative psychology and draws the notion of relational arity from formal systems in logic and computer science, including relational database theory. It formalizes task difficulty in terms of the number of entity and relations that must be simultaneously processed.

These frameworks form the basis for our task characterization approach, which we apply to \lkgb as an illustration. The \LKGBf was developed to address the lack of scalable evaluation tools for LLMs targeting KG tasks such as RDF serialization, SPARQL query generation and structured extraction. The framework supports a wide range of tasks with built-in correction cycles and output validation, emphasizing automated, reproducible evaluation across a broad selection of models. Here, we provide a short overview and description of the task groups used in \lkgb, for more details, see \cite{Meyer2023DevelopingScalableBenchmark,Frey2023Turtle,Frey2024AssessingEvolutionLLM,Meyer2024AssessingSparqlCapabilititesLLM,Meyer2025LLMKGBench3,Meyer2025EvaluatingLLMsForRdf}. 

\paragraph{RDF-related Tasks}

\begin{description}[font=\normalfont\itshape, style=multiline, leftmargin=46mm]
    \item[FactExtractStatic] Extract facts from a textual fact sheet and create a KG
    \cite{Meyer2023DevelopingScalableBenchmark,Frey2023Turtle}.
    \item[RdfConnectionExplainStatic] Find the shortest connection between two nodes in an RDF graph \cite{Frey2023Turtle,Meyer2025LLMKGBench3}.
    \item[RdfFriendCount] Identify the node with the most incoming edges \cite{Frey2023Turtle,Meyer2025LLMKGBench3}.
    \item[RdfSyntaxFixList] Correct a syntactically invalid RDF graph \cite{Meyer2025LLMKGBench3}.
    \item[TurtleSampleGeneration] Generate small Turtle KGs satisfying given requirements \cite{Meyer2023DevelopingScalableBenchmark,Frey2023Turtle}.
\end{description}

\paragraph{SPARQL-related Tasks}
\begin{description}[font=\normalfont\itshape, style=multiline, leftmargin=46mm]
    \item[Sparql2AnswerList] Given a small KG and a \sparqlselquery, return the respective result set for the query \cite{Meyer2024AssessingSparqlCapabilititesLLM}.
    \item[Text2AnswerList] Return the result set answering a given textual question on a given KG (withot a \sparqlselquery) \cite{Meyer2024AssessingSparqlCapabilititesLLM}.
    \item[Text2SparqlList] Given a KG and its description, construct a \sparqlselquery corresponding to a given natural language query \cite{Meyer2024AssessingSparqlCapabilititesLLM}.
    \item[SparqlSyntaxFixingList] Given a \sparqlselquery with syntax errors, return a corrected query \cite{Meyer2024AssessingSparqlCapabilititesLLM}.
\end{description}

\begin{table}
    \centering
    \caption{Overview of cognitive complexity frameworks and possible values.}
    \label{tab:taskCharacterizationValues}
    \begin{tabular}{ll}
    \toprule
       Framework  & Values \\
    \midrule
      Bloom's Taxonomy - Cognitive Processes~\cite{bloom1956}   & Remember, Understand, Apply, Analyze, Evaluate, Create \\
      Knowledge Dimensions~\cite{anderson2001}  & Factual, Conceptual, Procedural, Metacognitive \\
      Relational Complexity~\cite{halford1998}  & Low, Medium, High \\
      \bottomrule
    \end{tabular}
\end{table}

\section{Task Characterization}
To understand what kinds of abilities and operations are required by benchmarking tasks, we apply structured characterization criteria drawn from the three established frameworks, as introduced above. While we adopt terminology from cognitive psychology, we do not claim that LLMs engage in these processes in a human sense. Rather, we assess the extent to which their outputs reflect behavior consistent with such operations. Table~\ref{tab:taskCharacterizationValues} provides an overview of all possible values across the three frameworks. In the following, we describe the interpretation and assignment criteria for each value.

\paragraph{Bloom's Taxonomy - Cognitive Processes}
\begin{description}[style=multiline, leftmargin=25mm, font=\normalfont\itshape, parsep=0pt, itemsep=1pt]
    \item[Remember] The task depends primarily on ``mechanically'' recalling facts or definitions without further processing. 
     \item[Understand] The task requires interpreting given information, structures or queries without fundamentally transforming or generating new representations.  \item[Apply] A known procedure or pattern must be correctly executed, such as retrieval or following syntactic rules. 
     \item[Analyze] The task demands recognizing or decomposing relationships between data, especially when multiple elements or steps must be coordinated. \item[Evaluate] A task involves judging the correctness, relevance or quality of a result. 
     \item[Create] The task involves generating new content, such as generating queries or data structures.
\end{description}

\paragraph{Knowledge Dimensions}
\begin{description}[style=multiline, leftmargin=25mm, font=\normalfont\itshape, parsep=0pt, itemsep=1pt]
    \item[Factual] Task execution success depends on recalling or recognizing specific terminology, syntax elements or concrete information. 
    \item[Conceptual] Structural or relational understanding is necessary, such as schema structure, data models or logical organization. 
    \item[Procedural] The task requires a correct application of known methods, routines or transformation steps. 
    \item[Metacognitive] Awareness and control over one's strategies and thinking processes, such as selecting appropriate approaches, planning task execution or monitoring correctness, which might be relevant for more complex or interactive settings.
\end{description}

\paragraph{Relational Complexity}
\begin{description}[style=multiline, leftmargin=25mm, font=\normalfont\itshape, parsep=0pt, itemsep=1pt]
    \item[Low] The task involves interpreting or manipulating individual binary relations or isolated, simple structures with minimal dependencies. 
    \item[Medium] Multiple relations and entities must be processed simultaneously, such as coordinating several triples or variables in a query. 
    \item[High] The task involves multiple interrelated entities or nested dependencies that must be simultaneously considered, often requiring more abstract or hierarchical reasoning.
\end{description}

\begin{table}
    \centering
        \caption{Characterization of Benchmark Evaluation Tasks, first submitted at \cite{Meyer2025EvaluatingLLMsForRdf}}
    \label{tab:taskCharacterizationApplied}
    \begin{tabular}{ll lll}
    	\toprule
    	& Task & Cognitive Process & Knowledge Dimension & Relational Complexity \\
    	\midrule
    	\multicolumn{2}{l}{RDF related:} & & & \\
    	& FactExtractStatic & Understand, Create & Conceptual, Procedural & Medium \\
    	& RdfConnectionExplainStatic & Understand, Analyze & Conceptual & Medium \\	
    	& RdfFriendCount & Apply & Procedural & Low \\	
    	& RdfSyntaxFixList & Understand, Apply & Factual, Procedural & Low \\	
    	& TurtleSampleGeneration & Understand, Create & Conceptual, Procedural & Medium \\	
        \midrule
        \multicolumn{2}{l}{SPARQL related:} & & & \\
    	& Sparql2AnswerList & Understand, Apply & Conceptual, Procedural & Low \\	
    	& Text2AnswerList & Understand, Apply & Conceptual, Procedural & Low \\	
    	& Text2SparqlList & Understand, Create & Conceptual, Procedural & Low \\	
    	& SparqlSyntaxFixingList & Understand, Apply & Factual, Procedural & Low \\	
    	\bottomrule
    \end{tabular}\\
\end{table}

\subsection{Application to Benchmark Tasks}
The assigned values for each task are displayed in Table~\ref{tab:taskCharacterizationApplied}. Note that not all values across the frameworks are represented, as the current set of tasks does not span the full theoretical space. The assigned values represent the \emph{minimal} operational and structural requirements. Some variability in the relational complexity dimension is certainly possible given a prompt that requires more complex operations. 

We can observe several recurring value combinations. Most tasks fall into a characterization combining \emph{Understand} and \emph{Apply} as cognitive processes with \emph{Conceptual} and \emph{Procedural} knowledge dimensions and a \emph{Low} level of relational complexity. This reflects the prevalence of tasks requiring interpretation and rule application without substantial structural coordination. Tasks involving generation (\emph{FactExtractStatic}, \emph{TurtleSampleGeneration} and \emph{Text2SparqlList}) are naturally the only ones annotated with the \emph{Create} process. Among them, only the RDF-based generation tasks are assigned \emph{Medium} relational complexity, reflecting the need to coordinate multiple entities and their relationships when constructing a graph. In contrast, SPARQL generation tasks tend to result in single triple pattern and are thereby assigned \emph{Low} relational complexity.

A consistent, expected dependency can be observed between some processes and knowledge types. \emph{Factual} and \emph{Conceptual} knowledge always coincide with \emph{Understand}, as interpreting a meaning inherently involves factual or structural knowledge. On the other hand, \emph{Procedural} knowledge always coincides with \emph{Apply}, \emph{Analyze} or \emph{Create}, since carrying out a certain procedure by definition requires knowing the necessary steps. In the current task set, \emph{Factual} and \emph{Conceptual} knowledge do not co-occur, distinguishing between surface-level terminology and deeper structural comprehension. Similarly, \emph{Apply}, \emph{Analyze} and \emph{Create} do not co-occur, as they all describe mutually exclusive operations that either follow a procedure, decompose a structure, or generate new ones.

\section{Discussion and Outlook}
In this paper we proposed a task characterization that provides a complementary perspective on benchmark design and evaluation beyond accuracy metrics, inspired by theories of cognitive complexity. This can guide the creation of more balanced and targeted benchmarks by ensuring diversity across the different dimensions. Moreover, it enables identification of potential blind spots in model behavior for tasks that require similar processing.

We demonstrate how to assign the characterization values on a set of evaluation tasks from the \LKGBf. Several values do not appear in the task set due to current design preferences, but that does not imply that such dimensions are irrelevant or unassignable. In cognitive processes, we note \emph{Remember} which would describe a task that asks for reproduction of terminology or exact syntax, e.g., listing reserved SPARQL keywords from memory, while \emph{Evaluate} would require making judgments between alternative options, e.g., selecting the most efficient query. One knowledge dimension was not assigned, \emph{Metacognitive} knowledge, which might be tackled by tasks that require justification, such as explaining the reasoning behind a generated query. Finally, \emph{High} relational complexity would emerge in tasks requiring coordination of more than two entity roles simultaneously, like multi-dimensional event data or nested dependencies. This suggests a direction for extending task design to capture a broader range of structural demands.

The proposed framework could be applied to other benchmarks in the semantic web and beyond, allowing for cross-benchmark comparisons of task complexity profiles. It could also be integrated into such evaluation pipelines, helping understand the types of processes the models succeed or struggle with. In turn, this could support more systematic error analysis, design, and task selection.


\begin{acknowledgments}
This work was partially supported by grants from the German Federal Ministry of Education and Research (BMBF) to the projects ScaleTrust (16DTM312D) and KupferDigital2 (13XP5230L).
\end{acknowledgments}

\section*{Declaration on Generative AI}
The authors have not employed any Generative AI tools.

\bibliography{bibliography}

\clearpage
\pagestyle{plain}
\cfoot*{} 

\section*{Metadata for this article}

\begin{description}
    \item[Title:] Characterizing Knowledge Graph Tasks in LLM Benchmarks Using Cognitive Complexity Frameworks
    \item[Authors:] Sara Todorovikj \& Lars-Peter Meyer \& Michael Martin
    \item[original publication:] in proceedings of \href{https://2025-eu.semantics.cc/page/posters}{SEMANTiCS 2025 Poster Track},
      3.-5.9.2025 in Vienna, Austria
    \item[submitted for review:] 4.~7.~2025
    \item[peer review status:] accepted by peer review (4.~8.~2025)
    \item[submitted for publication:] 15.~8.~2025
    \item[publication date:] about 2025
    \item[Bibtex entry:] 
\end{description}

\begin{scriptsize}
  \begin{verbatim}
@InProceedings{Todorovikj2025CharacterizingKnowledgeGraphTasks,
 author    = {Todorovikj, Sara and Meyer, Lars-Peter and Martin, Michael},
 booktitle = {Proceedings of SEMANTiCS 2025 Poster Track},
 title     = {Characterizing Knowledge Graph Tasks in {LLM} Benchmarks Using Cognitive Complexity Frameworks},
 year      = {2025},
}  
  \end{verbatim}
\end{scriptsize}

\end{document}